\documentclass[10pt,twocolumn,letterpaper]{article}

\usepackage[listofformat=parens, justification=centering, subrefformat=subparens]{subfig}
\usepackage{cvpr}
\usepackage{times}
\usepackage{epsfig}
\usepackage{graphicx}
\usepackage{amsmath}
\usepackage{amssymb}
\usepackage{textcomp}
\usepackage{gensymb}

\renewcommand\cap[3]{\caption[#2]{\label{#1}\textsc{#2}. \small\textit{#3}}}
\usepackage[pagebackref=true,breaklinks=true,letterpaper=true,colorlinks,bookmarks=false]{hyperref}

\cvprfinalcopy 
\long\def\comment#1{}
\def\cvprPaperID{44} 

\ifcvprfinal\fi
\begin{document}

\newcommand\todo[1]{\textcolor{red}{TODO: \emph{#1}}}

\title{PARAPH: Presentation Attack Rejection by Analyzing Polarization Hypotheses}
\author{Ethan M. Rudd, Manuel G\"unther, and Terrance E. Boult\\
University of Colorado at Colorado Springs\\
Vision and Security Technology (VAST) Lab\\
\{erudd,mgunther,tboult\}@vast.uccs.edu
}

\maketitle
\thispagestyle{empty}

\begin{abstract}
For applications such as airport border control, biometric technologies that can process many capture subjects quickly, efficiently, with weak supervision, and with minimal discomfort are desirable.
Facial recognition is particularly appealing because it is minimally invasive yet offers relatively good recognition performance.
Unfortunately, the combination of weak supervision and minimal invasiveness makes even highly accurate facial recognition systems susceptible to spoofing via presentation attacks.
Thus, there is great demand for an effective and low cost system capable of rejecting such attacks.
To this end we introduce PARAPH -- a novel hardware extension that exploits different measurements of light polarization to yield an image space in which presentation media are readily discernible from Bona Fide facial characteristics.
The PARAPH system is inexpensive with an added cost of less than 10 US dollars.
The system makes two polarization measurements in rapid succession, allowing them to be approximately pixel-aligned, with a frame rate limited by the camera, not the system.
There are no moving parts above the molecular level, due to the efficient use of twisted nematic liquid crystals.
We present evaluation images using three presentation attack media next to an actual face -- high quality photos on glossy and matte paper and a video of the face on an LCD. In each case, the actual face in the image generated by PARAPH is structurally discernible from the presentations, which appear either as noise (print attacks) or saturated images (replay attacks).
\end{abstract}


\section{Introduction}
\label{sec:introduction}

\begin{figure}
{\small
\vspace{-1cm}
\begin{center}
\includegraphics[width=1.0\columnwidth]{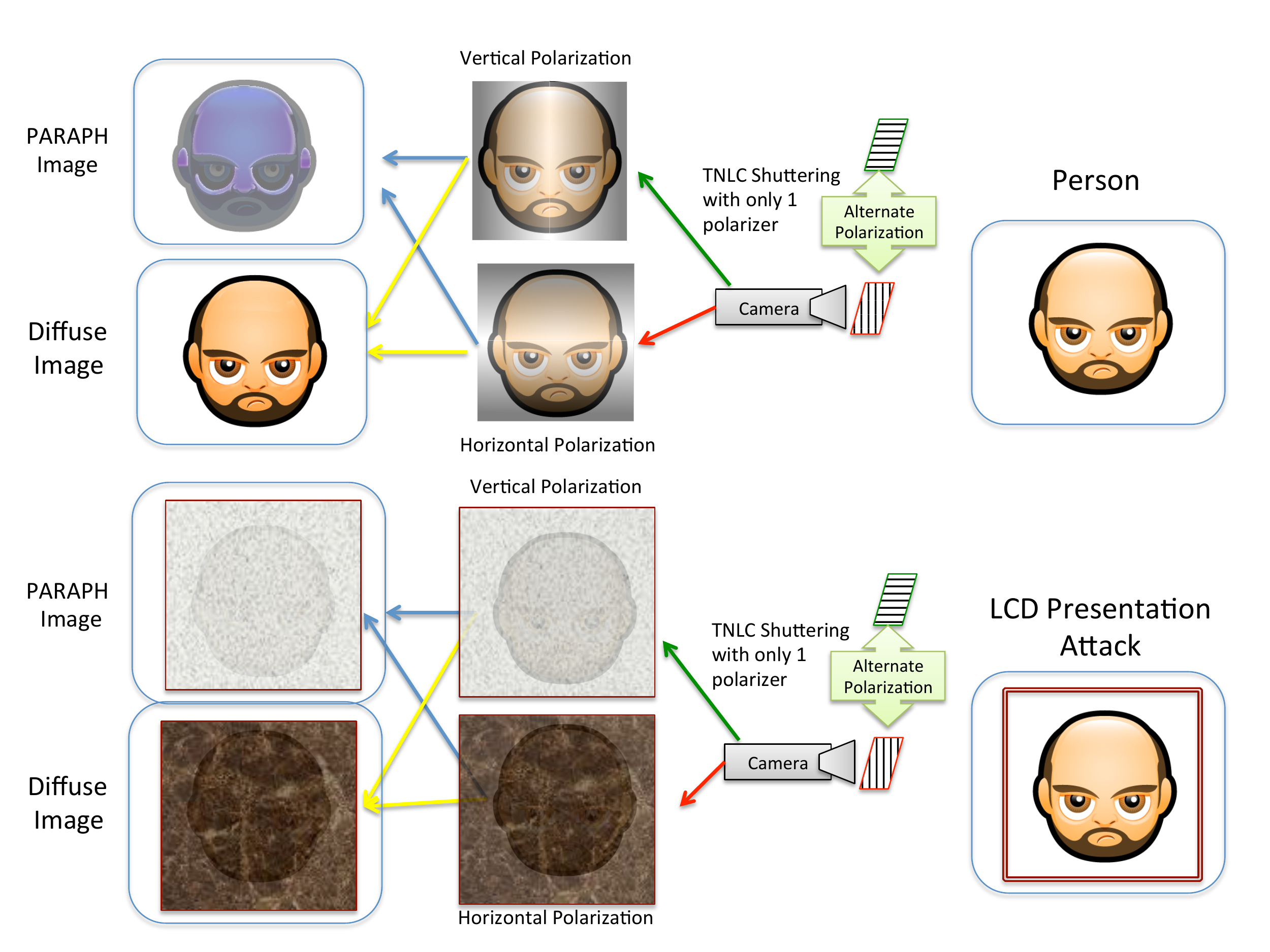}
\end{center}
}
\vspace{-0.5cm}
\cap{fig:teaser}{Conceptual schematic of PARAPH}{The system captures images under alternating horizontal and vertical polarizations, shuttering via a twisted nematic liquid crystal (TNLC).  These alternating images allow us to estimate a PARAPH image by taking the normalized per-pixel difference of a specular image and a diffuse image with reduced specular reflections. For a Bona Fide facial characteristic, the PARAPH image will have lots of structure related to facial geometry and the diffuse image can support the normal biometric system. When a presentation attack with an LCD or display is imaged, the entire screen will be polarized, the PARAPH image will lack face structure, and the diffuse image will be mostly noise.
}
\vspace{-.5cm}
\end{figure}

Face is an appealing biometric modality because it is more efficient and less invasive than other modalities such as fingerprint and iris.
Automatic face recognition has been researched for several decades, and in some respects has been shown to surpass human face recognition capabilities \cite{otoole2007surpass}.
\comment{In 2007, O'Toole et al. \cite{otoole2007surpass} demonstrated that automatic face recognition systems were able to outperform humans for face verification tasks -- ascertaining whether or not two images contained the same identity -- on a dataset of cooperative subjects under typical indoor imaging conditions.
Thus, in applications in which imaging conditions can be reasonably controlled, e.g., border control systems, autonomous systems should be able to replace the human in the loop.}
However, there is still one large problem that prevents the use of fully autonomous face recognition systems for security-critical access control applications: namely, many face recognition algorithms can easily be spoofed by \textit{presentation attacks} \cite{chingovska2014spoofing}.

A presentation attack is formally defined as a ``presentation to the biometric data capture subsystem with the goal of interfering with the operation of the biometric system''~\cite{ISOPDF1}.
Such attacks pose a challenge to all biometric systems, but particularly for the face modality it is \textit{very} easy for an attacker to acquire high-quality facial image or video data.
Moreover, resolution demands for such facial presentation attacks are modest, and high-quality printing or electronic display can produce an image that, when captured by a face recognition system, is nearly identical to the original image.
In addition, high-quality portable displays, in the form of phones, tablets and laptops, often make presenting facial images/videos straightforward.
Therefore, the production of face spoofing images, formally referred to as \textit{artefact} images~\cite{ISOPDF1}, is well within the technological reach of billions of people.
To date, most techniques used in face anti-spoofing attempt to analyze the original content of the image or video to detect artifacts that were introduced by a printer or a video compression algorithm.
Only few algorithms incorporate additional information by using infra-red (IR) or near-infrared (NIR) imagery for presentation attack detection \cite{galbally2014antispoofing}.
While IR has its merits, IR imaging introduces noticeable costs, and spatial resolution is inherently poorer due to both longer wavelength and focal plane array limitations \cite{short2015exploiting}.
In this paper, we seek a lower cost means of augmenting currently deployed visible wavelength cameras to reject presentation attacks.

To this end we introduce PARAPH, a system that delineates Bona Fide facial characteristics from spoof media via light polarization analysis.
A conceptual schematic of the approach is shown in Fig.~\ref{fig:teaser}.  Because skin polarizes reflected light perpendicular to the surface normal and polarizes the diffuse component in the plane of the normal, the normalized difference of horizontal and vertical polarization components, the \textit{PARAPH image}, is tightly tied to facial geometry. Thus, a human capture subject will elicit a large response for legitimate facial structure, whereas an artefact in presentation media will elicit little to no response.
While the PARAPH image can be used to detect and reject presentations, standard facial recognition algorithms can be applied to the diffusely polarized component. Using the diffuse component may actually improve recognition performance by removing many specularities.

The analysis of polarization itself is not new to computer vision or  biometrics. In computer vision applications, polarization analysis under \textit{passive illuminations} was pioneered by Wolff and Boult \cite{wolff1991constraining}, who demonstrated how to use camera-based polarization analysis to constrain surface normals, estimate material properties, and discriminate edge types (e.g., occluding vs. albedo).
Polarization measurements are used in several biometric sensors \cite{rowe2005multispectral,rowe2007multispectral,wildes1997iris}, although those almost exclusively rely on crossed or circular polarizers to manage illumination and specular reflections under \textit{active illumination}.
For facial recognition applications, polarization analysis has recently been applied to enhance recognition between long and medium wave infrared (LWIR and MWIR) probe images and the visible spectrum by fusing histogram of oriented gradient (HOG) features over several Stokes images~\cite{short2015exploiting}.

In this paper we show, both theoretically and empirically, that simple polarization analysis can be used to discriminate Bona Fide face presentation from attack presentations, i.e., facial photos/video displayed on media including prints, LCD, LED, and AMOLED displays.
Our design leverages a linear polarizer and a fast-switching twisted nematic liquid crystal to serve as an analyzer at two polarization angles, in opposition to more traditional/complex linear polarization analysis \cite{wolff1995polarization,wolff1991constraining}.
Unlike stereo or thermal/IR based approaches, incorporating this design into a camera introduces a materials cost of less than 10 US dollars, even with our simple prototype; a cost which could be dramatically reduced in mass-production.

\section{Theoretical Foundation of Polarization}
\label{sec:background}

Light behaves as a transverse wave, with electric and magnetic field components oscillating orthogonally about the direction of propagation, the Poynting vector.
The orientation of the electric field is known as the \textit{polarization} of light.
A more detailed overview of polarization can be found elsewhere \cite{pedrotti1993introduction}, but we shall introduce the subject matter in sufficient detail to motivate PARAPH.

When light encounters a surface, an electromagnetic interaction occurs, which depends on the polarization, wavelength, and phase of the electromagnetic waves.
Often, an exchange of energy causes a change in wavelength of the light (color), but a phase shift can also occur -- for example, when light is reflected, the phase changes by $180\degree$.
Depending on a material's properties, e.g., the direction of freest flow of electrons, light of certain polarizations can pass through the material easily, while light of other polarizations is reflected or absorbed.
The transmitted light is \textit{partially polarized} with one dominant polarization.
Materials that allow  one polarization to be almost purely transmitted while blocking the orthogonal polarization are generally referred to as \textit{polarizers}.
In this paper, we shall predominantly constrain the discussion to \textit{linear polarizations}, where the orientation of the electromagnetic field remains fixed over time.

The intensity of light transmitted through a linear polarizer depends on the relative angle between polarizer and light polarization, according to Malus' law:
%
\begin{equation}
I = I_0\,\cos^2(\theta),
\end{equation}
where $I_0$ is the intensity of purely polarized incident light, $I$ is the intensity of the transmitted light, and $\theta$ is the relative angle between incident light polarization and polarizer orientation. \textit{Unpolarized} light refers to light with no preference for polarization -- or approximately equal polarizations from all angles. Such light is commonly emitted from a radiating source like a lamp or the sun. The expected intensity of unpolarized light that gets through a linear polarizer will therefore be half the incident intensity because:
\begin{equation}
I = I_0 \frac{1}{\pi} \int_0^{\pi} \cos^2(\theta)\,d\theta = \frac{1}{2} I_0.
\end{equation}

\textit{Partially polarized} light consists of a superposition of purely polarized light and unpolarized light. As a polarizer is rotated, the transmitted light will vary with $\cos^2(\theta)$ from a minimum intensity $I_{min}$ to a maximum intensity $I_{max}$, where $I_{min} = \frac{1}{2} I_0$ corresponds to the unpolarized portion of the light, when no polarized light gets through ($\theta$ is a multiple of $\frac{\pi}{2}$), and $I_{max} - I_{min}$ is the intensity of the purely polarized light. This phenomenon is commonly referred to as the \textit{transmitted radiance sinusoid}.

When light interacts with a surface, it becomes partially polarized, depending on the surface composition.
Reflected light tends to be polarized parallel to the plane of incidence -- the plane containing the Poynting vector and surface normal -- (s-polarization), while transmitted light tends to be polarized parallel to the incident plane (p-polarization).
The s-polarized light reflected directly off the surface is a glossy \textit{specular reflection}, while light that passes into the surface and internally reflects several times before passing back out is known as \textit{diffuse reflection}.
Although diffuse reflection illuminates the surface, it is \textit{generally} dull and unpolarized due to many interactions with planar facets in the sub-surface.
However, an exception occurs from diffuse reflections under extreme angles of incidence, e.g., on occluding contours, when almost all light propagating to the observer is multiply-internally reflected along the occluding edge before being p-polarized from the output transmission.
This results in a subtle aura-like effect around the edges of an object, which has a polarization orthogonal to the specularly reflected component.
Note that this partial polarization of the diffuse ``reflection''  is actually a result of transmission.

\comment{
  The amount of light of a given polarization that gets reflected/absorbed or transmitted at a given angle depends on material properties which are characterized by indices of refraction.
  One of the most common and straightforward relationships to this end is \textit{Brewster's angle}, which characterizes the angle of incidence at which p-polarized light will be completely transmitted or absorbed. This angle is given by $\theta_B = arctan\left(\frac{n_1}{n_2}\right)$, where $n_1$ and $n_2$ are the respective indices of refraction of the incident and transmitted materials.}

While an in-depth quantitative treatment of the polarization effects of different materials is well beyond the scope of this paper, the important point is that polarization characteristics are highly material and texture dependent. This property of polarized light leads us to hypothesize that we can discriminate between presentation attack media and legitimate faces by examining differences in their polarization signatures, an approach that we refer to as PARAPH.

To estimate the transmitted radiance sinusoid and analyze the polarization of a surface requires at least three, and often uses four different polarization measurements per pixel.
Early work on polarimetric vision mechanically rotated a polarizer between subsequent frames, but rotating fast enough to achieve video rate polarization imaging is complex and hence not cheap; \href{http://bossanovatech.com}{bossanovatech.com} sells cameras using computer-controlled rotating polarizers. Alternatively, one can use beam-splitting and multiple cameras, such as in the system commercially available from \href{http://fluxdata.com}{fluxdata.com}, but this too is expensive. An even higher-end approach, available from \href{http://4dtechnology.com}{4dtechnology.com}, is a single high-speed camera with a grid of pixel sized polarizers, precisely aligned to replace the traditional Bayer pattern. Such cameras can support hundreds of frames per second and can work in NIR, but the cost per unit starts at 10,000 and increases to over 25,000 US dollars. A much lower cost ``do it yourself'' approach for obtaining full polarization imaging  using two different types of polarization cameras and a Raspberry Pi2 can be found online.\footnote{{\interfootnotelinepenalty1000000000\url{http://www.diyphysics.com/category/instrumentation/polarimetric-imaging}}}
While all of these approaches allow full linear polarization (Stokes) state estimation, which takes at least three measurements, for presentation attack rejection -- at least of basic attacks -- we believe we can significantly reduce complexity and cost using only partial state estimation. Thus  we explore an approach using only two polarization measurements.

\section{Discerning Presentation Attacks}
\label{sec:approach}

\begin{figure}[t!]
  \centering
  \includegraphics[width=\linewidth]{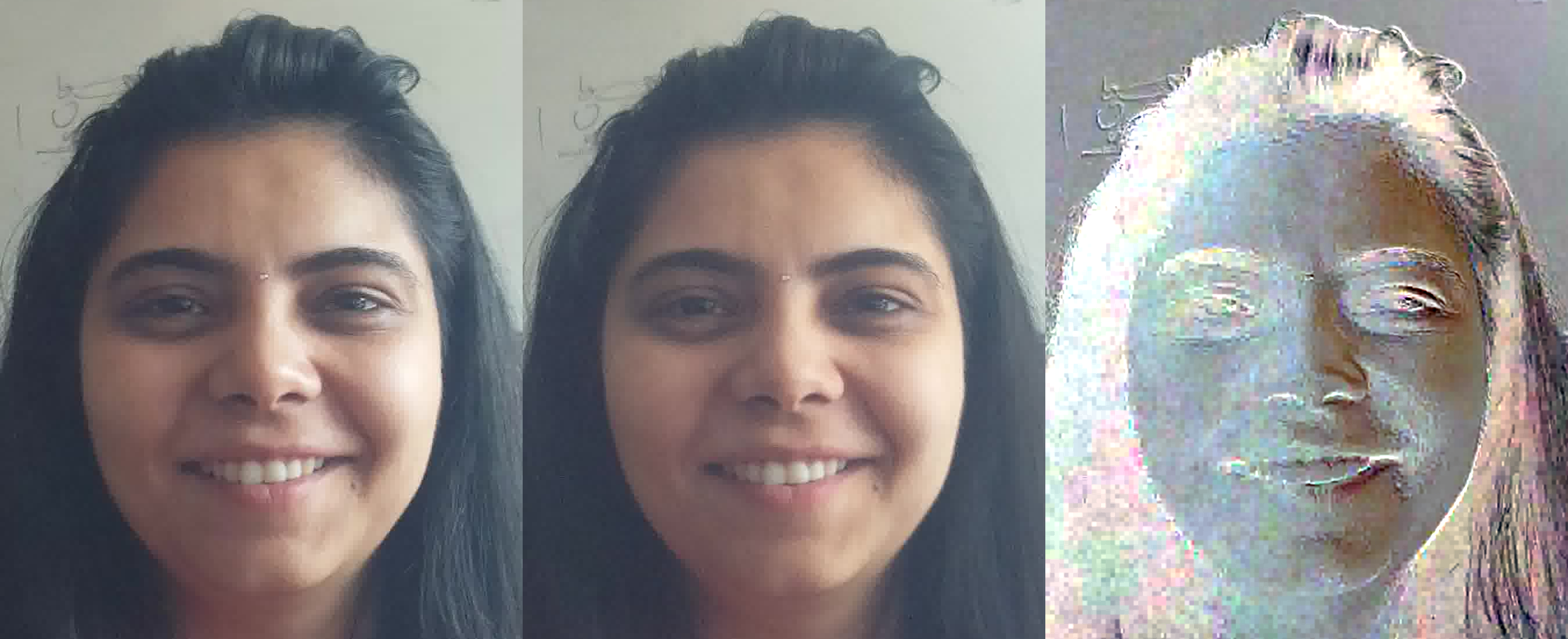}
  \cap{fig:svati_polar}{Polarization of a face}{Images were obtained by manually rotating a polarizer in front of the camera of a Samsung Galaxy S6 Edge\texttrademark{} smartphone. Left: linear horizontal polarization. Center: linear vertical polarization. Right: PARAPH image $I_P$ from Eq.~\eqref{eq:paraph_img}. Under horizontal polarization, the intensity of specularly polarized light is noticeably greater. Note that scaling has been applied for visualization.
}
\end{figure}

While we could simply examine faces through arbitrary polarizations and observe the optical effects, our goal is to develop a simplified, but still principled approach, which clearly differentiates legitimate faces from spoof media.
From the discussion in Sec.~\ref{sec:background} and our knowledge about facial geometry, we make the following observations:

\begin{enumerate}
\setlength{\parskip}{0pt}
\setlength{\itemsep}{0pt}
\item From the vertically elongated geometry of human faces, we would expect p-polarization from diffuse ``reflection'' to be maximized (on average) at a polarization angle of 0\degree{} (vertical) on the sides of the face and at an angle of 90\degree{} (horizontal) on top and chin.

\item We expect that specularly reflected light from the cheeks, nose, and forehead should be polarized at 90\degree{} since, by definition, the visible portion of the face is \textit{facing} the viewing plane.

\item Because the orientation for $I_{\mathrm{max}}$ will be directly related to surface normals of the face, the polarization image will tightly match face geometry.

\end{enumerate}

From these observations, let $I_h$ be an image taken under a 90\degree{} polarization, and $I_v$ be an image taken under a 0\degree{} polarization. Let us further assume that the images are pixel-aligned. Then the normalized image of maximum contrast in intensity due to polarization, the \emph{PARAPH} image $I_P$, will approximately be:
\vspace{-0.5em}
\begin{equation}
\label{eq:paraph_img}
\frac{I_{\mathrm{max}}-I_{\mathrm{min}}}{I_{\mathrm{max}}+I_{\mathrm{min}}} \approxeq I_P = \frac{I_h - I_v}{I_h + I_v},
\vspace{-0.5em}
\end{equation}
which is the average maximum change in amplitude of the transmitted radiance sinusoid for either vertically or horizontally polarized light. An example of $I_P$ is shown in Fig.~\ref{fig:svati_polar} along with respective $I_v$ and $I_h$ images. Face images will have noticeable differences under $I_v$ and $I_h$. The intuition discussed in Sec.~\ref{sec:background} and previous research~\cite{wolff1991constraining} suggest that faces and presentation materials will have much different $I_P$ images.
$I_P$ is trivial to obtain, though the constraint that $I_v$ and $I_h$ be pixel-wise aligned can make low-cost systems a bit more complex.  To create a low-cost system capable of acquiring a PARAPH image with a single camera and no moving parts at scales above the molecular level, we utilize \textit{twisted nematic liquid crystals} (TNLCs) with their remarkable ability to twist light polarizations.

TNLC molecules naturally align along their elongated edges. When contained in an enclosure with a grating of parallel nanometer-thick ridges on both ends -- each end oriented orthogonally to the other -- the end-molecules ``get stuck'' in the ridges, and the bound molecules assume a helical chain, which has the electromagnetic effect of a phase-retarder, \textit{twisting} the polarization of the emitted light by 90\degree. However, when an electric potential of sufficient strength is applied across the two ends of the polarizer, the resulting electric force \textit{breaks the molecular bond}, and causes the molecules to realign, axially oriented perpendicular to both ridge gratings. The polarization of the light that passes through no longer changes.
Placing a TNLC in front of a vertical polarizer leads to the following effect: when no potential is applied to the TNLC, incident horizontally polarized light will adopt a vertical polarization after passing through the liquid crystal and will pass unhindered through the vertical polarizer. Incident vertically polarized light on the TNLC will be emitted from the liquid crystal with a horizontal polarization and will not pass through the vertical polarizer.
When an electric potential is applied, the opposite happens: Vertical light incident on the liquid crystal will maintain its original orientation after passing through the TNLC, and will pass freely through the polarizer, while horizontal light incident on the TNLC will remain horizontally polarized and, thus, not pass through the polarizer. By placing a lead on each side of the TNLC and toggling current, a camera sensitive only to light intensity, placed in front of the polarizer (which is placed in front of the liquid crystal), can easily measure the intensity of two orthogonal components of polarization, $I_v$ and $I_h$.

Provided that the refresh rate of both the camera and liquid crystal is greater than the rate of noticeable motion, an approximate pixel-wise alignment can be achieved. Modern webcams achieve frame rates of 30 FPS or greater -- even the inexpensive ones -- but what about liquid crystals? Conveniently, TNLCs are precisely the technology used to toggle active-shutter glasses for 3D televisions, typically operating at 120 Hz, which is more than enough to obtain subsequent frames of orthogonal light polarizations (incident on the TNLC) for a 30 FPS camera.

For our design, we obtained both liquid crystal and polarizer by disassembling a pair of \textit{G7 Universal} active-shutter 120 Hz kids 3D glasses with \textit{Duo Sync Technology}\texttrademark, which we bought for 9 US dollars on eBay.
We used a 5V battery DC power supply soldering wires to a lead on each side of the TNLC, with simple switching to toggle power on/off.  The net added cost needed to measure polarization (excluding stabilizing clamps, tripod, and an inexpensive webcam) was less than 10 US dollars. Even with this inexpensive prototype, the differences between legitimate faces and a variety of presentation attacks can be easily distinguished, which we will show in the next section.

\section{Evaluation and Discussion}
\label{sec:experiments}

Unfortunately, we cannot readily compare our algorithm to other state-of-the-art face anti-spoofing systems that are evaluated on default benchmark datasets \cite{zhang2012database,chingovska2012effectiveness}, since we need polarization information from the live data. Consequently, the images and video streams in these datasets are insufficient for our purposes.
Instead, we conducted our own small-scale tests using high quality printed images on matte paper, high quality printed images on glossy paper, and videos taken from LCD monitors.

To demonstrate that our system works even under non-trivial lighting with low resolution, we used 352x288 images taken from a 10 US dollar \textit{VAlinks\textregistered} USB 2.0 webcam.
Prior to computing the PARAPH image, we convolved the $I_h$ and $I_v$ images with a 5x5 Gaussian filter to denoise the images and reduce sensitivity to misalignments.
Comparisons of PARAPH images $I_P$ along with the raw $I_h$ and $I_v$ images are shown in Figs.~\ref{fig:svati_spoofs} and ~\ref{fig:steve_spoofs}.
Although PARAPH images are computed in color, they are shown in grayscale to enhance visibility.
A uniform scaling/shifting to a visible pixel space is also applied, since $I_P$ in Eq.~\eqref{eq:paraph_img} can only assume values between $-1$ and $+1$.
Fig.~\ref{fig:svati_spoofs} shows the subject of the same identity as shown in the printed images, while Fig.~\ref{fig:steve_spoofs} shows a subject of different identity.
As expected, each of the presentation media yields a PARAPH image that is noticeably different than that of an actual face.

\begin{figure*}[!ht]
  \centering
  \subfloat[Live face vs. face displayed on an LCD\label{fig:lcd_color_svati}]{\includegraphics[width=\linewidth]{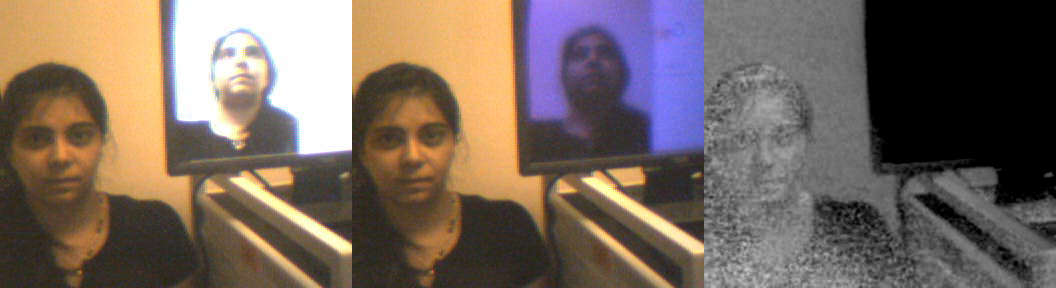}}\\
  \subfloat[Live face vs. face printed on glossy paper\label{fig:printed_glossy_svati}]{\includegraphics[width=\linewidth]{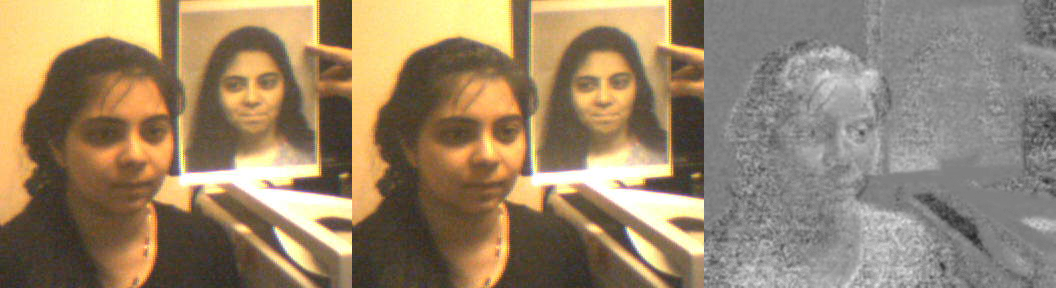}}\\
  \subfloat[Live face vs. face printed on matte paper\label{fig:printed_no_gloss_svati}]{\includegraphics[width=\linewidth]{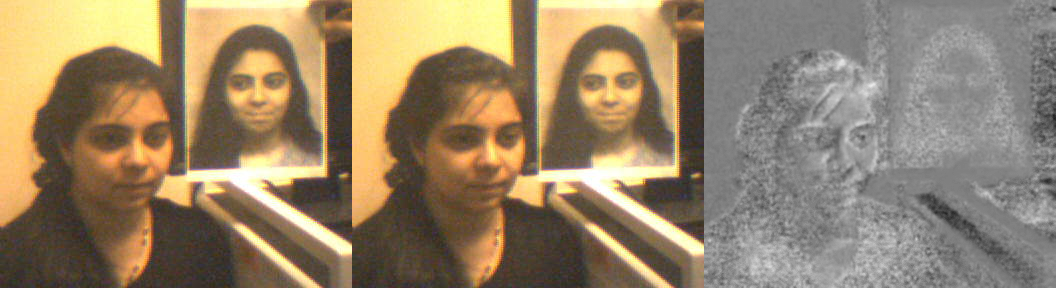}}\\[1ex]
  \cap{fig:svati_spoofs}{Faces: real and presentation media subject 1}{Vertical polarization images (left), horizontal polarization images (center), and PARAPH images (right) are shown for each of the faces. As a function of the vertical polarization of the LCD in \protect\subref*{fig:lcd_color_svati}, we can clearly differentiate this attack even with one polarizer. Note that the intensity of specular reflection on the face is greater for images taken at horizontal polarizations than for vertical polarizations, but as an inherent property of the screen the intensity of the LCD is far greater for vertical polarizations, and thus yields a very low intensity PARAPH image. While the high quality glossy print in \protect\subref*{fig:printed_glossy_svati} may be good enough to spoof facial recognition systems that use conventional cameras, its PARAPH image looks little like a face due to a relatively uniform polarization of the gloss. 
A high quality printed photo on matte paper has a PARAPH image that traces the silhouette of a face as shown in \protect\subref*{fig:printed_no_gloss_svati}, but is devoid of fine structure. 
With the cheap webcam that we used, the dark regions of the image have noticeable noise which results in artificially large values for Eq.~\ref{eq:paraph_img} which leads to the apparent face silhouette.
Future work better noise reduction.
Note: scaling has been applied for visualization.}
\end{figure*}

\begin{figure*}[!ht]
  \centering
  \subfloat[Live face vs. face displayed on an LCD\label{fig:lcd_color_steve}]{\includegraphics[width=\linewidth]{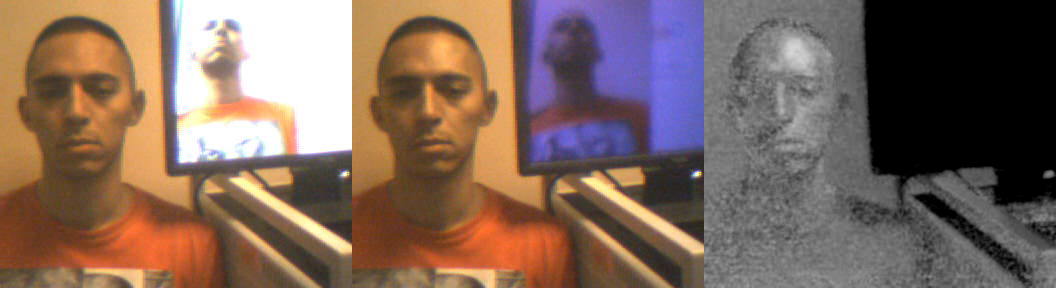}}\\
  \subfloat[Live face vs. face printed on glossy paper\label{fig:printed_no_gloss_steve}]{\includegraphics[width=\linewidth]{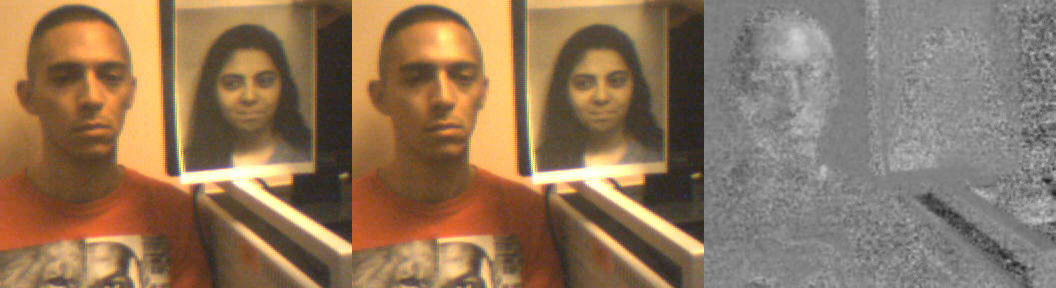}}\\
  \subfloat[Live face vs. face printed on matte paper\label{fig:printed_glossy_steve}]{\includegraphics[width=\linewidth]{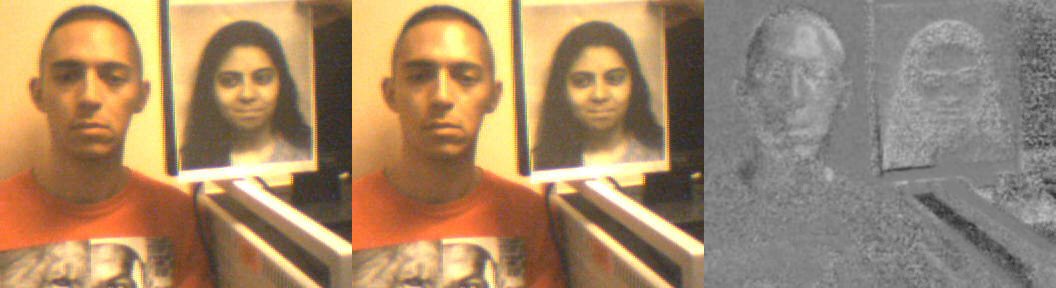}}\\[1ex]
  \cap{fig:steve_spoofs}{Faces: real and presentation media subject 2}{Vertical polarization images (left), Horizontal polarization images (center), and PARAPH images (right) for a second subject. Note: scaling has been applied for visualization.}
\end{figure*}

For each medium, the differences between the presentation image and the actual face are rather extreme.
The intensity from the LCD screen is very low. We cannot even discern an image, because LCD screens themselves emit linearly polarized light; crossed polarizers -- one in front of and one behind the LCD -- ordinarily yield a black screen.
Thin film transistors cause the liquid crystals in the pixels of interest to align and let light through.
The refresh rate (60 Hz in this case) refers to the rate of toggle for each row of pixels.
From Fig.~\subref{fig:lcd_color_svati} and Eq.~\eqref{eq:paraph_img}, we can see that the light emitted from the LCD is vertically oriented. If the LCD screen were non-vertically polarized, it would be visible, but no facial-like structure would be prevalent because light is emitted; not reflected in a single polarization.

Glossy photo paper is often perceived as \textit{higher quality} by humans than conventional printer paper.
Interestingly, though both presentation media are clearly differentiable from an actual face, the glossy paper is more clearly differentiable.
The matte print resolution is the same, but a noisy silhouette of the face is far more noticeable.
This effect could be caused by polarization differences from ink around planar facets of the matte paper, which do not occur in a smooth gloss coating. 
However, none of the PARAPH images show the fine-grained structure in presentation media that is clearly visible for the actual faces.

To attain non-trivial illumination conditions, subjects were placed in the corner of a room with a wall on the left side as seen by the viewer, \textit{which attenuates specular reflection} from that side of the face.
In the PARAPH image, we can clearly see much more granularity and far less structure on the wall side of the face, resulting from less specular and more diffuse polarization.
The wall side of the face is also darker than the unpolarized background, while the opposite side of the face is brighter.
The brightness is caused by specular reflection from the horizontal polarizer on the right side of the face, i.e., $\mathrm{sign}(I_P) > 0$ because $I_h > I_v$. Since the diffuse component (vertical polarization) is subtracted in Eq.~\ref{eq:paraph_img}, and specular reflection on the left side of the face is more or less absent due to the non-specular wall, the left side of the face in $I_P$ is negative and therefore much darker than the right side and slightly darker than the background.
While a diffuse component exists on the right side of the face as well, this component is dwarfed by the relatively bright specular reflection in $I_h$. Thus, $I_P$ for this region assumes positive values.
As explained in Sec.~\ref{sec:background}, the polarizations of diffuse and specular components differ because the diffuse ``reflection'' is is caused by transmission of internally reflected light along occluding contours.
This explains the intensity differences between sides of the face in our PARAPH images. Under uniform lighting conditions, the face appears symmetric, consisting of either dark ``diffuse'' reflection or bright specular reflection, but in either case, structural information for the face is apparent.
Our analysis has shown that even under uneven and non-trivial illumination conditions with an inexpensive noisy low-resolution camera, the PARAPH image of the Bona Fide face is readily distinguishable from the presentation spoofs.

\section{Other Approaches to Face Anti-Spoofing}
\label{sec:related}

Although other approaches to anti-spoofing differ dramatically from our PARAPH approach, in that none of them leverage passive light polarization analysis, we believe that this other research bears mentioning. A broader survey on face spoofing and antispoofing techniques in particular is presented by Galbally et al.~\cite{galbally2014antispoofing}.

Most researchers attempt to perform face antispoofing based on the original image or video data to counter printed and video-based replay attacks.
These approaches have the advantage that they do not require any special hardware and they integrate nicely into existing image- or video-based face recognition systems \cite{chingovska2013joint}.
To counter printed attacks, where a photograph of the victim is printed on paper and held in front of the camera, texture based algorithms have been used to detect artifacts that were introduced by the printer.
Motion-based approaches try to find a difference in motion between the foreground (face) and the scene context \cite{anjos2013motion}, or they require the cooperation of the subject by prompting for a specific head movement, e.g., nodding, smiling, or blinking \cite{galbally2014antispoofing,kollreider2008liveness}.
However, these systems have the disadvantage that they require a video-based recognition system. They cannot effectively be used in static systems that provide only a single image for verification \cite{galbally2014antispoofing}.

Texture and motion based algorithms have achieved relatively high spoofing detection rates \cite{chingovska2013competition}, but most of them are limited to anticipated attack types.
They are not robust to \textit{novel types} of spoofing attacks, such as 3D mask attacks \cite{erdogmus2014masks}, for which yet a different set of algorithms has been developed \cite{erdogmus2013kinect,kose2013reflectance}, nor are they resistant to spoofing with makeup.\footnote{Spoofing a face recognition system by makeup won the ICB 2013 TABULA RASA Spoofing Challenge, see \url{http://www.tabularasa-euproject.org/evaluations/tabula-rasa-spoofing-challenge-2013}}
Additionally, research suggests that texture- and motion-based antispoofing algorithms are highly dependent on the dataset, on which they were trained, and are not yet ready to be applied in production \cite{pereira2013realworld}.

To date, only a few other works have attempted to detect spoof attempts by leveraging specialized hardware to acquire information other than raw 2D image intensity data.
An obvious attempt to detect print and replay attacks, which are usually displayed on a flat surface, is to use 3D imaging techniques.
However, to the best of our knowledge, no such counter-measure has yet been proposed; the closest approach that we could find uses a light field camera \cite{kim2014lightfield}.
Other systems, which require active illumination of the scene with LEDs to delineate attacks from real accesses based on reflectance information, have been presented \cite{kim2009radiance,zhang2011multispectral}, but these works have only demonstrated their capabilities of detecting 3D masks made of silicon or paper.
Another approach uses thermal imaging \cite{sun2011tir} for liveness detection, capturing both IR and visible spectrum images at the same time.
However this approach requires good cross-spectral spatial alignment, which the authors performed using hand-annotated eye and mouth coordinates.
The system's capabilities under autonomous cross-spectral alignment were not tested.
Further, no study to our knowledge has been performed on how thermal presentation attack detection systems are affected by environmental conditions (e.g., cold weather, direct sunshine).

The PARAPH technique that we present in this paper is different than all the other existing techniques in that -- in principle -- it can detect any kind of currently known spoofing attack.
How to effectively detect print or replay attacks using polarization has been demonstrated in Sec. \ref{sec:experiments}.
Mask or prosthetic attacks can theoretically be detected by classifying the difference in polarization profiles between the mask/prosthetic and human skin \cite{wolff1991constraining}, but the technique presented in this paper would likely need to be refined and extensive experiments conducted -- we did not have access to 3D printed masks at the time of this writing.
Two of the biggest advantages of our approach over other specialized hardware techniques are first, that it is completely passive, neither requiring active illumination of the scene nor extra/specialized infrared cameras, and second, that it is quite cheap by comparison.
PARAPH also does not prompt the capture subject for explicit facial cues.
Hence, we believe PARAPH could easily and inexpensively be incorporated into existing facial recognition systems for authentication and access control, e.g., border control, building access, and electronic payment, and that with additional R\&D could be extended to passive surveillance applications as well.


\section{Conclusion}
\label{sec:conclusion}

PARAPH is a novel passive low-cost approach, which, in principle, defeats many known and possibly several unanticipated presentation attack types.
Unlike purely algorithmic solutions, many of which rely on unreliable or active behavioral cues, PARAPH is based on robust laws of physics. The system is difficult to spoof because polarization depends on both shape and material type of the media in question.
Our approach is not the only presentation attack detection approach to exploit the laws of physics through hardware, but it is far less expensive than others.
We have not yet tested the system using masks, prosthetics, and makeup.
However -- theoretically speaking -- our approach should be able to detect these attacks since the attack media have different polarization characteristics than human skin, but to work well it may require higher quality imaging and a full Stokes state vector. We leave this evaluation to future work.

There are several ways in which PARAPH could be extended.
First, higher quality materials could be used.
Second, active lighting of known angle and polarization could be applied to enhance the quality of PARAPH images.
Third, polarization measurements could be added to obtain a full Stokes state vector.
Finally, polarization information could be fused with other antispoofing techniques, e.g., we could exploit cross-spectrum (visible, SWIR, MWIR, LWIR) polarization information or perform stereoscopic polarization measurements incorporating depth information. While all of these techniques could enhance quality and difficulty to spoof, they come at the cost of additional complexity and expense. Weighing these tradeoffs is an important subject, which we leave to future research, along with extended experimentation and the collection of a polarimetric dataset of capture subjects and presentation attacks.


{\small
\bibliographystyle{ieee}
\bibliography{paper}

\begin{thebibliography}{10}\itemsep=-1pt

\bibitem{anjos2013motion}
A.~Anjos, M.~M. Chakka, and S.~Marcel.
\newblock Motion-based counter-measures to photo attacks in face recognition.
\newblock {\em IET Biometrics}, 2013.

\bibitem{chingovska2012effectiveness}
I.~Chingovska, A.~Anjos, and S.~Marcel.
\newblock On the effectiveness of local binary patterns in face anti-spoofing.
\newblock In {\em BIOSIG}, 2012.

\bibitem{chingovska2013joint}
I.~Chingovska, A.~Anjos, and S.~Marcel.
\newblock Anti-spoofing in action: joint operation with a verification system.
\newblock In {\em CVPR}, 2013.

\bibitem{chingovska2014spoofing}
I.~Chingovska, A.~Anjos, and S.~Marcel.
\newblock Biometrics evaluation under spoofing attacks.
\newblock {\em T-IFS}, 9(12):2264--2276, 2014.

\bibitem{chingovska2013competition}
I.~Chingovska et~al.
\newblock The 2nd competition on counter measures to {2D} face spoofing
  attacks.
\newblock In {\em ICB}, 2013.

\bibitem{pereira2013realworld}
T.~de~Freitas~Pereira, A.~Anjos, J.~M. De~Martino, and S.~Marcel.
\newblock Can face anti-spoofing countermeasures work in a real world scenario?
\newblock In {\em ICB}, 2013.

\bibitem{erdogmus2013kinect}
N.~Erdogmus and S.~Marcel.
\newblock Spoofing in {2D} face recognition with {3D} masks and anti-spoofing
  with kinect.
\newblock In {\em BTAS}, 2013.

\bibitem{erdogmus2014masks}
N.~Erdogmus and S.~Marcel.
\newblock Spoofing face recognition with {3D} masks.
\newblock {\em T-IFS}, 2014.

\bibitem{galbally2014antispoofing}
J.~Galbally, S.~Marcel, and J.~Fierrez.
\newblock Biometric antispoofing methods: A survey in face recognition.
\newblock {\em IEEE Access}, 2:1530--1552, 2014.

\bibitem{ISOPDF1}
ISO/IEC.
\newblock Information technology -- biometric presentation attack detection --
  part 1: Framework.
\newblock ISO/IEC 30107-1, International Organization for Standardization /
  International Electrotechnical Commission, Geneva, Switzerland, 2016.

\bibitem{kim2014lightfield}
S.~Kim, Y.~Ban, and S.~Lee.
\newblock Face liveness detection using a light field camera.
\newblock {\em Sensors}, 14(12), 2014.

\bibitem{kim2009radiance}
Y.~Kim, J.~Na, S.~Yoon, and J.~Yi.
\newblock Masked fake face detection using radiance measurements.
\newblock {\em JOSA-A}, 26(4):760--766, 2009.

\bibitem{kollreider2008liveness}
K.~Kollreider, H.~Fronthaler, and J.~Bigun.
\newblock Verifying liveness by multiple experts in face biometrics.
\newblock In {\em CVPR Workshops}, 2008.

\bibitem{kose2013reflectance}
N.~Kose and J.~L. Dugelay.
\newblock Reflectance analysis based countermeasure technique to detect face
  mask attacks.
\newblock In {\em DSP}, 2013.

\bibitem{otoole2007surpass}
A.~J. O'Toole, P.~J. Phillips, F.~Jiang, J.~H. Ayyad, N.~Penard, and H.~Abdi.
\newblock Face recognition algorithms surpass humans matching faces over
  changes in illumination.
\newblock {\em TPAMI}, 29(9):1642--1646, 2007.

\bibitem{pedrotti1993introduction}
F.~L. Pedrotti, L.~S. Pedrotti, and L.~M. Pedrotti.
\newblock {\em Introduction to optics}.
\newblock Pearson Education, 1993.

\bibitem{rowe2005multispectral}
R.~K. Rowe, S.~P. Corcoran, K.~A. Nixon, and R.~E. Ostrom.
\newblock Multispectral imaging for biometrics.
\newblock In {\em Biomedical Optics}, 2005.

\bibitem{rowe2007multispectral}
R.~K. Rowe, U.~Uludag, M.~Demirkus, S.~Parthasaradhi, and A.~K. Jain.
\newblock A multispectral whole-hand biometric authentication system.
\newblock In {\em Biometrics Symposium}, 2007.

\bibitem{short2015exploiting}
N.~Short, S.~Hu, P.~Gurram, and K.~Gurton.
\newblock Exploiting polarization-state information for cross-spectrum face
  recognition.
\newblock In {\em BTAS}, 2015.

\bibitem{sun2011tir}
L.~Sun, W.~Huang, and M.~Wu.
\newblock {TIR/VIS} correlation for liveness detection in face recognition.
\newblock In {\em CAIP}, pages 114--121, 2011.

\bibitem{wildes1997iris}
R.~P. Wildes.
\newblock Iris recognition: an emerging biometric technology.
\newblock {\em Proceedings of the IEEE}, 85(9):1348--1363, 1997.

\bibitem{wolff1995polarization}
L.~B. Wolff and A.~G. Andreou.
\newblock Polarization camera sensors.
\newblock {\em IVC}, 13(6):497--510, 1995.

\bibitem{wolff1991constraining}
L.~B. Wolff and T.~E. Boult.
\newblock Constraining object features using a polarization reflectance model.
\newblock {\em TPAMI}, (7):635--657, 1991.

\bibitem{zhang2012database}
Z.~Zhang, J.~Yan, S.~Liu, Z.~Lei, D.~Yi, and S.~Z. Li.
\newblock A face antispoofing database with diverse attacks.
\newblock In {\em ICB}, 2012.

\bibitem{zhang2011multispectral}
Z.~Zhang, D.~Yi, Z.~Lei, and S.~Z. Li.
\newblock Face liveness detection by learning multispectral reflectance
  distributions.
\newblock In {\em FG}, pages 436--441, 2011.

\end{thebibliography}
}
\end{document}